\documentclass[sigconf]{acmart}
\usepackage{amsmath}
\usepackage{amsbsy}
\usepackage{graphicx}

\usepackage{enumitem}
\usepackage{booktabs}
\usepackage{subfigure}
\usepackage{subcaption}
\usepackage{mathrsfs}

\usepackage{xcolor}

\newcommand{\tensor}[1]{\boldsymbol{\mathscr{#1}}}

\AtBeginDocument{%
  }

\copyrightyear{2026}
\acmYear{2026}
\setcopyright{cc}
\setcctype{by}
\acmConference[KDD '26]{Proceedings of the 32nd ACM SIGKDD Conference on Knowledge Discovery and Data Mining V.2}{August 09--13, 2026}{Jeju Island, Republic of Korea}
\acmBooktitle{Proceedings of the 32nd ACM SIGKDD Conference on Knowledge Discovery and Data Mining V.2 (KDD '26), August 09--13, 2026, Jeju Island, Republic of Korea}
\acmDOI{10.1145/3770855.3819008}
\acmISBN{979-8-4007-2259-2/2026/08}

\begin{document}

\title{Tensor Methods: A Unified and Interpretable Approach for Material Design}


\author{Shaan Pakala}
\affiliation{%
  \institution{University of California, Riverside}
  \institution{Dept. of Computer Science \& Engineering}
  \city{Riverside, CA}
  \country{USA}
}
\email{spaka002@ucr.edu}

\author{Aldair E. Gongora}
\affiliation{%
  \institution{Lawrence Livermore National Laboratory}
  \institution{Materials Engineering Division}
  \city{Livermore, CA}
  \country{USA}
}
\email{gongora1@llnl.gov}

\author{Brian Giera}
\affiliation{%
  \institution{Lawrence Livermore National Laboratory}
  \institution{Data Science Institute}
  \city{Livermore, CA}
  \country{USA}
}
\email{giera1@llnl.gov}

\author{Evangelos E. Papalexakis}
\affiliation{%
  \institution{University of California, Riverside}
  \institution{Dept. of Computer Science \& Engineering}
  \city{Riverside, CA}
  \country{USA}
}
\email{epapalex@cs.ucr.edu}

\renewcommand{\shortauthors}{Shaan Pakala, Aldair E. Gongora, Brian Giera, \& Evangelos E. Papalexakis}

\begin{abstract}

When designing new materials, it is often necessary to tailor the material design to have some desired properties (e.g., an optimal Young's Modulus value). As the set of material design parameters grows, the search space grows exponentially, making the actual synthesis and evaluation of all combinations of designs virtually impossible. Even using traditional computational methods, such as Finite Element Analysis (FEA), becomes too computationally heavy to search this design space. Recent methods use machine learning (ML) surrogate models to more efficiently determine optimal material designs; unfortunately, these methods often (i) are notoriously difficult to interpret and (ii) under perform when the training data comes from a non-uniform sampling of the entire design space. In this work, we suggest the use of tensor completion methods as an all-in-one approach for interpretability and predictions. We observe that classical tensor methods are able to compete with traditional ML methods in predictions, with the added benefit of their interpretable tensor factors (which are given completely for free, as a result of the prediction). In our experiments, we are able to rediscover physical phenomena via the tensor factors, indicating that our predictions are aligned with the true underlying physics of the problem. This also means these tensor factors could be used by experimentalists to identify potentially novel patterns, given we are able to rediscover existing ones. We also study the effects of both types of surrogate models (traditional ML \& tensor-based) when we encounter training data from a non-uniform sampling of the design space. We observe some more specialized tensor methods that are able to give better generalization in these non-uniform sampling scenarios (e.g., neural tensor completion methods), due to the low-rank constraint. We find the best generalization comes from a tensor model, which is able to improve upon the baseline ML methods by up to 5\% on aggregate $R^2$, and halve the error in some out of distribution sections of the search space.

\end{abstract}

\ccsdesc[500]{Computing methodologies~Machine learning}
\ccsdesc[500]{Information systems~Data mining}
\ccsdesc[500]{Applied computing~Physical sciences and engineering}

\keywords{Tensor completion, material design, surrogate modeling, interpretability, data imbalance}


\maketitle

\renewcommand{\thefootnote}{}
\footnotetext{\textbf{Code}: \url{https://github.com/shaanpakala/Tensor-Methods-for-Material-Design}}
\renewcommand{\thefootnote}{\arabic{footnote}}

\section{Introduction}

\begin{figure*}[!t]
    \centering
    \begin{center}
        \includegraphics[width = 0.72\textwidth]{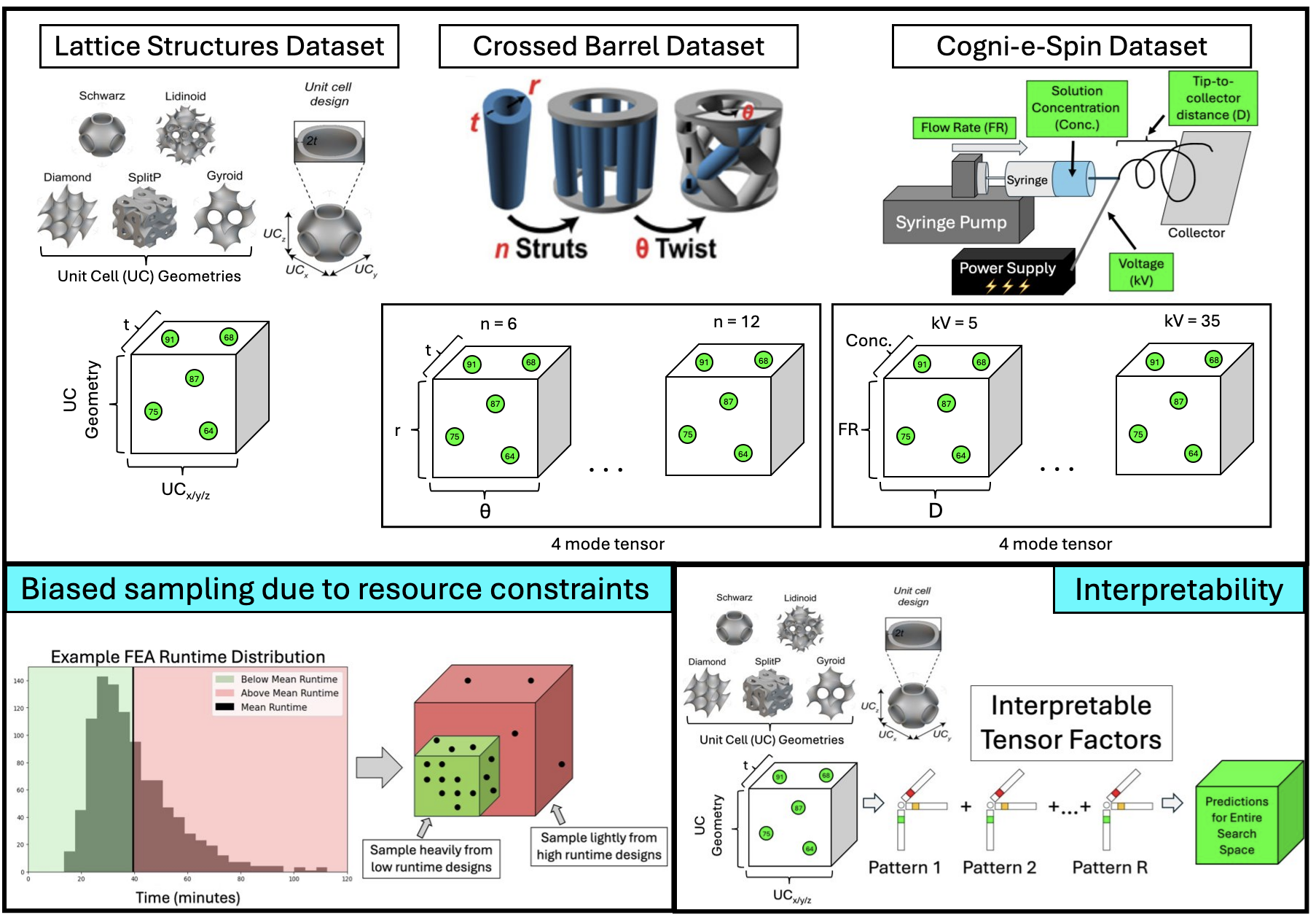}
        \caption{We present various material design problems as instances of tensor completion, in order to abstract away the various design components and use tensor methods to infer the entire search space. We first visualize how we are able to convert various material design optimization problems (designing an optimal lattice structure, 3D printed structures, and electrospinning configurations for nanofiber development) into instances of tensor completion problems. Then we depict a biased sampling of the design space (which is what a surrogate model's training data might consist of in practice), resulting from oversampling from low-cost designs and undersampling from high-cost designs (in this visualization "cost" is measured as FEA simulation runtimes). Furthermore, we display a Canonical Polyadic Decomposition (CPD) tensor model, which generates interpretable tensor factors automatically as a result of its predictions of the design space. Please note the histogram representing FEA simulation times is synthetic, and inspired by the actual distribution of FEA runtimes from the original paper \cite{gongora2024accelerating} (the real histogram can be found in the original paper's supplementary figures). The parts of the images depicting the actual lattice structure designs \cite{gongora2024accelerating} and the crossed barrel designs \cite{gongora2020bayesian} come from their original papers and are both CC-BY licensed. \label{overview_figure}}
    \end{center}
\end{figure*}

Designing specific structures or materials with desired properties (e.g. Young's modulus) becomes very challenging as new design variables are added \cite{jiao2021artificial, yeo2018materials}. The number of possible combinations to explore is exponential with respect to the number of design variables added, encouraging the use of computational methods to accelerate the navigation of this vast design space \cite{wu2021topology, pan2020design}. Traditional computational methods, such as Finite Element Analysis (FEA), can still prove to be very computationally heavy for exploring a large design space, which motivates the recent work on using machine learning (ML) surrogate models to more efficiently search the material design space \cite{alderete2022machine, challapalli2021machine, gongora2024accelerating, gongora2021using, liang2021benchmarking, pakala2025surrogate, pakala2025tensor}. There are several impactful applications of this ML-based surrogate modeling, including designing vehicles and aircrafts \cite{wang2021surrogate} and self-driving/autonomous laboratories \cite{gongora2020bayesian, snapp2023autonomous}.

Although there have been great advances with ML-based surrogate modeling for these types of problems, there is still much work to be done to address challenges in (i) interpretability and (ii) generalization when the data comes from a non-uniform sampling of the search space. 

(i) In terms of interpretability for traditional ML methods, they typically range from linear regression (very interpretable) to neural networks (not very interpretable). Unfortunately, linear regression suffers in prediction performance, and neural networks require an additional model to be trained to interpret, such as Shapley Additive Explanations (SHAP) \cite{NIPS2017_7062}, which can be time-consuming and expensive.

For this reason, we suggest the use of tensor completion methods for interpretable surrogate modeling in these material design problems. In this work we show that tensor completion methods are able to compete with traditional ML methods, with the extra benefit of producing interpretable tensor factors for free (as a result of the predictions). We find that tensor methods are also able to rediscover the underlying physical phenomena via the tensor factors, indicating that our predictions are aligned with the true physics of the problem. Not only is this great for model interpretation and being able to observe what exactly the model is learning (as opposed to a black-box ML model), but this can also be useful for domain experts to explore potentially novel patterns in the data, given we can rediscover existing ones. 

(ii) In addition to this, ML methods tend to overfit when the training data comes from a non-uniform sampling of the entire search space, which is very common in practice. This could happen when an experimentalist mainly designs materials that are convenient/cheaper to produce, instead of randomly selecting material designs to synthesize and analyze. This could also happen if the experimentalist synthesizes materials that they hypothesize to be optimal, which is again not uniformly randomly across the search space. Although these types of scenarios are very common in practice, there is very limited work studying this type of non-uniformly sampled training data for ML surrogate models. Some works attempt to address this data imbalance by using synthetic data \cite{zhang2026machine, sreedev2025leveraging, jiang2025review}, which would also require another model to be trained for the synthetic data.

In this work we study how different surrogate models might behave without requiring additional synthetic data in the out of distribution regions. We also study to what extent tensor methods may be useful in these scenarios, and observe their ability to better generalize to the out of distribution parts of the design space. We find the best generalization comes from a tensor model, which is able to improve upon the baseline ML methods by up to 5\% on aggregate $R^2$, and halve the error in some out of distribution sections of the search space.  

In this work we make several distinct contributions on surrogate modeling for designing optimal materials, mainly suggesting another tool (tensor completion methods) that experimentalists might find useful when designing optimal materials. Our contributions can be summarized as follows:
\begin{itemize}
    \item \textbf{Interpretability of tensor methods}: We study the interpretability of tensor methods and their ability to rediscover existing physical phenomena in the problem, and suggest they could be used in practice to help identify potentially novel phenomena
    \item \textbf{Non-uniform sampling for training data}: We investigate how ML surrogate models behave when the training data comes from a non-uniform sampling of the design space (which is very common in practice)
    \item \textbf{Tensor completion to handle non-uniform sampling}: We suggest tensor completion to handle some of these non-uniform sampling scenarios, and show cases where they can generalize better than traditional ML methods
    \item \textbf{Benchmarking tensor completion}: We also generally benchmark the utility of tensor completion as a surrogate model in these types of tasks
    \item \textbf{Variety of datasets}: Our experiments cover 3 material design scenarios: designing optimal lattice structures \cite{gongora2024accelerating}, designing optimal 3D printed structures with regards to toughness \cite{gongora2020bayesian}, and designing optimal electrospinning configurations to produce nanofibers \cite{mahdian2026cogni}
    \item \textbf{Public code}: Our code used for experiments is publicly available at \url{https://github.com/shaanpakala/Tensor-Methods-for-Material-Design}.
\end{itemize}
\section{Methods}

Here we give a detailed description on the methods used in this work. This includes our pipeline for tensor-based surrogate modeling, its training, and the datasets and metrics used for evaluation.

\subsection{Tensor Decomposition for Interpretable Surrogate Modeling}

In Figure \ref{overview_figure}, we show the complete surrogate modeling pipeline we are proposing. After experimentally determining the performance for various material structures, we can convert those observations to entries of a tensor and use tensor completion to infer the performance of the remainder of the design search space.

\subsubsection{Tensors \& Tensor Decomposition}

Tensors are a general term for multidimensional arrays. In other words, a 1st order tensor is just a vector, and a 2nd order tensor is a matrix. In this work, we will be looking at higher order tensors, denoted $\tensor{X}$. Tensor decomposition is a general term for expressing a tensor as several smaller factors, and is an extension of matrix factorization to multidimensional datasets. Tensor decomposition is a powerful tool, not only for data compression, but also for analyzing multidimensional datasets. For example, the smaller tensor factors are commonly used for pattern discovery in these multidimensional datasets. A common form of tensor decomposition is the Canonical Polyadic Decomposition (CPD)  
\cite{kolda2009tensor,sidiropoulos2016tensor}. CPD expresses a tensor as a sum of rank-one tensors. A rank $R$ CPD of a third-order tensor $\tensor{X} \in \mathbb{R}^{I \times J \times K}$ would be expressed as:

\[
\tensor{X} \approx \sum_{r=1}^{R} (\mathbf{a}_r \circ \mathbf{b}_r \circ \mathbf{c}_r)
\]

where $\circ$ denotes outer product, $\mathbf{a}_r \in \mathbb{R}^I, \mathbf{b}_r \in \mathbb{R}^J, \text{ and } \mathbf{c}_r \in \mathbb{R}^K$.

\subsubsection{Interpreting the Tensor Factors} In the CPD each set of rank 1 components can be attributed to a different pattern in the dataset. A common example is decomposing a 2-dimensional movie-users matrix (where each entry is the rating a user gives to a movie) into rank 1 factors \cite{koren2009matrix}. Here each rank 1 component might correspond to a movie genre (e.g. comedy, horror), and the rank 1 component corresponding to each movie and user ($\mathbf{a}_r$ and $\mathbf{b}_r$ from the previous notation) shows the expression of each pattern (genre) in each movie or user, based on the component's magnitude.

\subsubsection{Tensor Completion} Furthermore, tensor decomposition can be used for tensor completion, which is the process of filling in the missing values of a tensor. For a CPD model, once the tensor factors are estimated (from the sparse tensor), the missing values can be predicted using the sum (across all ranks $r\in R$) of the products of the corresponding indices of the rank 1 components. A prediction for tensor entry $\tensor{X}_{i, j, k}$ would simply be $\sum_{r=1}^{R} (\mathbf{a}_{r,i} \times \mathbf{b}_{r,j} \times \mathbf{c}_{r,k})$, where $\mathbf{a}_{r,i}$ is the $i$-th entry of the $r$-th rank 1 component $\mathbf{a}_r \in \mathbb{R}^I$.

\subsubsection{Tensor Models}

There are several classes of tensor completion methods. There are classical methods, relying solely on the tensor factors, such as CPD \cite{kolda2009tensor}, CPD-S (a CPD-based method that imposes smoothness constraints on some of the tensor factors) \cite{ahn2021time, 10825934}, and TuckER \cite{balazevic2019tucker}. There are also more advanced neural tensor completion methods such as NeAT \cite{ahnneural} or CoSTCo \cite{liu2019costco}, which leverage the use of neural networks in addition to the tensor factors. The classical tensor models are able to produce interpretable tensor factors for free, as a result of the tensor completion. The neural tensor methods tend to be less interpretable due to their reliance upon neural networks.

For our experiments, we use CPD, CPD-S, NeAT, and CoSTCo. For our CoSTCo implementation, we randomly generate several sets of initial tensor factors and aggregate these using a Convolutional Neural Network (CNN) for better stability, since it is significantly more sensitive to the initialization.

\subsubsection{Tensor Completion Training}

For training our tensor completion models, we first randomly initialize the factor matrices to generate an initial (random) reconstruction of the full tensor. Then we perform gradient descent using the error of the observed tensor values and the associated predictions, to update our parameters (tensor factor matrices) to get a more accurate reconstruction of the full tensor. For our error function, we use Mean Squared Error (MSE) = $\frac{1}{n}\sum_{i=1}^{n}(y_i - \hat y_i)^2$. To optimize the factor matrices (and neural network parameters for neural tensor models) with regards to the error term, we use Adam \cite{kingma2014adam}.

\subsection{Evaluation}

Here we outline how we conduct our evaluation on our surrogate models, including datasets, different settings on these datasets, baseline comparisons, and evaluation metrics.

\subsubsection{Datasets}

We briefly describe the datasets used in this work, as well as the design space in Figure \ref{datasets_table}. A more detailed explanation of the datasets can be found in the datasets' original papers. Visualizations of the datasets and their respective design variables can be found in Figure \ref{overview_figure}.

\begin{enumerate}
    \item \textbf{Lattice Structures Dataset} \cite{gongora2024accelerating}: Contains data for lattice structure designs and their corresponding Young's Modulus (E) and ratio of Young's Modulus to mass ($\tilde{E}$), which are experimentally determined. These lattice structure designs are solely parameterized by the Unit Cell ($UC$), including the $UC$ geometry, $UC$ thickness, and $UC_X$, $UC_Y$, $UC_Z$ lengths. This dataset was generated using FEA simulations.

    \item \textbf{Crossed Barrel Dataset} \cite{gongora2020bayesian}: Contains data for 3D printed structures, parameterized by number of struts (n), strut thickness (t), strut radius (r), and the angle at which the structure is twisted ($\theta$). The outcome of interest is the mechanical toughness of the material. This dataset was generated from experimentally 3D printing and evaluating each material design, using an automated robotic system.

    \item \textbf{Cogni-e-Spin Dataset} \cite{mahdian2026cogni}: Contains data for electrospinning parameters and their resulting nanofiber morphologies. This electrospinning process is generally parameterized by Solution Concentration, Voltage ($kV$), Flow Rate ($\frac{mL}{h})$, Tip-to-Collector Distance ($cm$), and the Polymer. To characterize the nanofiber morphology, this dataset considers the resulting Fiber Diameter ($nm$). This dataset was generated by synthesizing the data from several other studies.

\end{enumerate}

\begin{figure*}[!h]
\centering
\small
\setlength{\tabcolsep}{4pt}
\begin{tabular}{c|ccccc|c|cc}
\multicolumn{1}{c}{Dataset} & \multicolumn{5}{c}{Design parameters (\# of values)} & \multicolumn{1}{c}{Outcome(s)} & \multicolumn{1}{c}{Total} & \multicolumn{1}{c}{Observed} \\
\toprule
Lattice Structures \cite{gongora2024accelerating} & $UC$ Geometry (5) & $UC$ Thickness $t$ (2) & $UC_X$ (3) & $UC_Y$ (3) & $UC_Z$ (3) & $E$ \& $\tilde{E}$ & 270 & 270\\ 
Crossed Barrel \cite{gongora2020bayesian} & $n_{\text{struts}}$ (4) & Strut $\theta$ (9) & Radius $r$ (11) \ & Thickness $t$ (3) & - & Toughness & 1188 & 600 \\ 
Cogni-e-Spin \cite{mahdian2026cogni} & Solution Conc. (35) & Voltage (23) & Flow Rate (22) & Tip Dist. (22) & Polymer (3) & Fiber Diameter & 1,168,860 & 288 \\
\bottomrule
\end{tabular}
\caption{The design spaces and outcomes of interest for the datasets used in our experiments. We also display the total possible design combinations (Total) and the number of observed combinations (Observed) in each dataset, from the set of design parameter values. For the lattice structures dataset, design are solely parameterized by the unit cell (UC): Geometry, Thickness, and the length in the X, Y, and Z directions. For the crossed barrel dataset, the designs are solely parameterized by the struts: number of struts, $\theta$ for the struts' twist, strut radius, and strut thickness. For the Cogni-e-Spin dataset, designs are parameterized by the electrospinning configurations: solution concentration, voltage, flow rate, tip-to-collector distance, and the polymer. \label{datasets_table}}
\end{figure*}

\subsubsection{Evaluation Scenarios}

We consider two main scenarios for evaluation. The first being a typical supervised learning scenario, where the training data is selected uniformly randomly from the entire set of data. The second being a biased sampling of the design space, as is typical in real-world scenarios. We simulate this biased sampling by drawing the training data (which corresponds to the experimentally validated data in practice) from two disjoint sets of the total data. One set represents a small subsection of the entire design space where training data is heavily sampled from, and another set for the remainder of the design space, from which training data is sampled very little. A visualization of our exact training sampling is available in Figure \ref{biased_sampling_heatmaps}.

\subsubsection{Baseline Comparisons}

For our baseline methods, we use Linear Regression (LR), Gaussian Process (GP), Random Forest (RF), XGBoost, CatBoost, and a Multi-Layer Perceptron (MLP). These were either recommended by each datasets' original paper (GP \& CatBoost), or generally solid ML baselines. These are all implemented using Scikit-Learn \cite{pedregosa2011scikit} (except for XGBoost \cite{chen2016xgboost} and CatBoost \cite{dorogush2018catboost}).

\subsubsection{Performance Metrics}

To quantify the regression performance, we use $R^2$ = $1- \frac{\sum(y_i - \hat{y_i})^2}{\sum(y_i-\bar{y})^2}$, Mean Absolute Error (MAE) = $\frac{1}{n}\sum_{i=1}^{n}|y_i - \hat{y_i}|$, Root Mean Squared Error (RMSE) = $\sqrt{\frac{1}{n}\sum_{i=1}^{n}(y_i - \hat{y_i})^2}$, and Mean Absolute Percentage Error (MAPE) = $\frac{1}{n}\sum_{i=1}^{n}|\frac{y_i - \hat{y_i}}{y_i}|$, for the observed value $y_i$ and corresponding predicted value $\hat{y_i}$.

To compare tensor factors, we use the Factor Match Score (FMS), implemented by TensorLy \cite{kossaifi2019tensorly}. To calculate the FMS, first TensorLy calculates the optimal permutation of the tensor factors (since they may not be arranged in the same order), and then the following FMS is computed (for a CPD):

\[
FMS = \sum_{r=1}^R \prod_{m=1}^M\frac{c_{m,r}^T \tilde{c}_{m,r}}{\|c_{m,r}^T\| \|\tilde{c}_{m,r}\|}
\]

where $c_{m,r}$ \& $\tilde{c}_{m,r}$ correspond to the $r$-th rank 1 component corresponding to tensor mode $m$, for two different decompositions. An FMS of 0 indicates no match and 1 indicates a perfect match.
\section{Experiments}

\subsection{Interpretability}

A typical method of interpreting a feature's importance in predictions is using Shapley Additive Explanations (SHAP) \cite{NIPS2017_7062}, which requires a post-hoc analysis of the trained ML model. A benefit of using linear tensor models such as CPD or CPD-S is the ability to interpret the tensor factors, which are given for free as a result of the tensor completion. Here we study the tensor factors to see if we can uncover some effects of the design parameters on the output of interest.

\subsubsection{Lattice structures}

\begin{figure}[!ht]
    \centering
    \includegraphics[width = 0.4\textwidth]{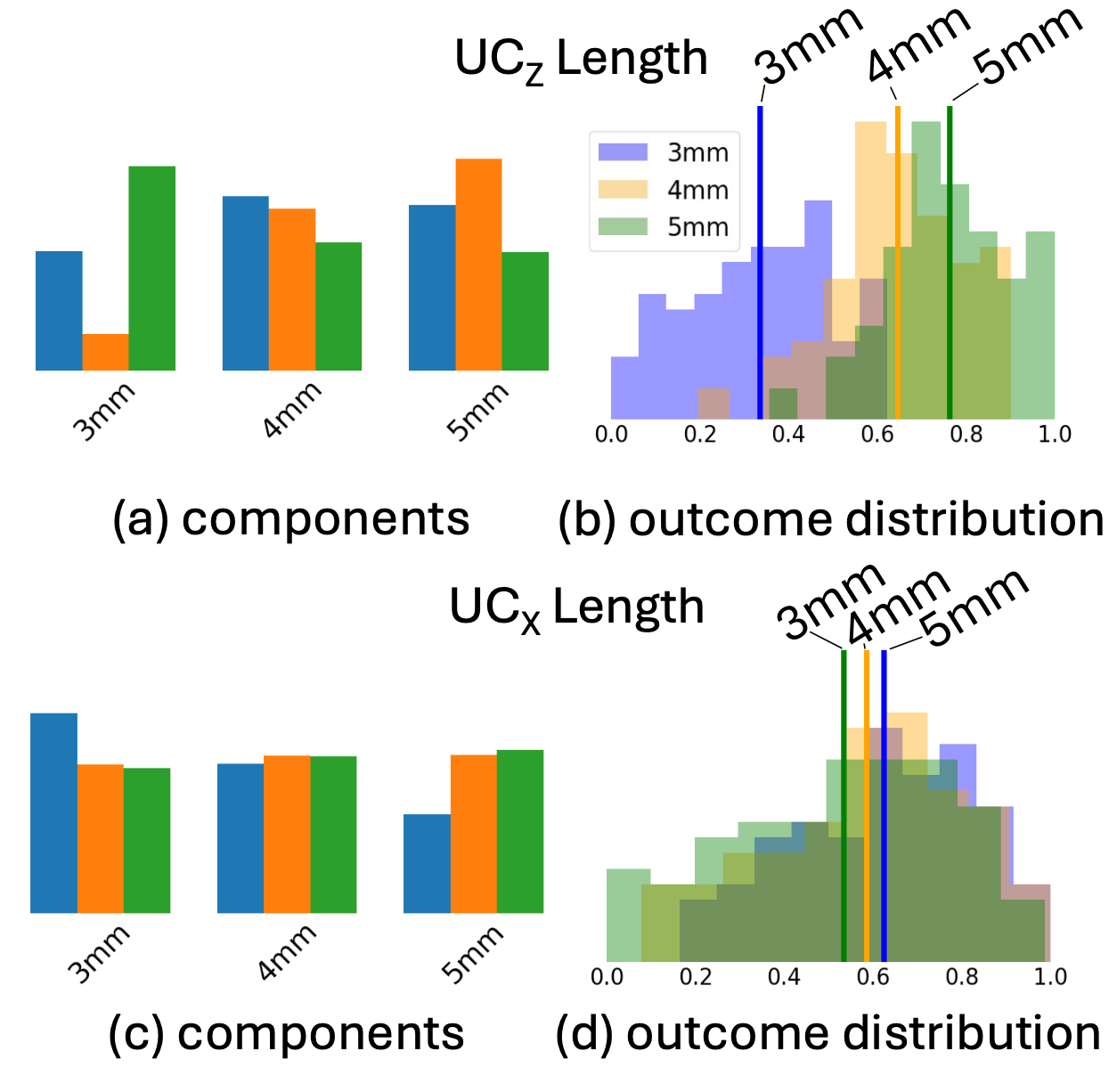}
    \caption{For the lattice structures dataset, we compute CPD (rank 3) factors using 25\% of training values and display the factors generated in (a) \& (c). Then we show the true distribution with respect to these design parameter values in (b) \& (d) (vertical lines for mean). For the $UC_Z$ length, we notice a distinct increase in component magnitude with an increasing length in (a). We can associate this with a correlation between the $UC_Z$ length and $\tilde{E}$, which we also observe from the true distribution in (b). This finding is also observed in the dataset's original paper \cite{gongora2024accelerating}. We compare this with the $UC_X$ length parameter in (c), which does not express a clear trend, as observed in the true distribution in (d) as well. \label{cpd_factors_viz_lattice}}
\end{figure}

First we consider the lattice structures dataset, since we get excellent prediction capabilities for this dataset. Also the dataset's original paper \cite{gongora2024accelerating} already has some work on surrogate model interpretation for this, allowing us to attempt to rediscover these interpretations. From a CPD (rank 3) tensor completion model, we plot the normalized (by $\ell_2$-norm) components for each mode (i.e. design parameter) in Figure \ref{cpd_factors_viz_lattice}.

From Figure \ref{cpd_factors_viz_lattice}, we observe a clear trend with the $UC_Z$ Length design parameter, where an increase in the length corresponds to an increase in the outcome, which is similarly found in the dataset's original paper (they use a GP with SHAP for interpretability) \cite{gongora2024accelerating}.

\textbf{Discussion}: The ability to rediscover this trend via the CPD factors allows us to have confidence that the model predictions are in agreement with the physics of the problem. Previous ML methods require the use of expensive external models such as SHAP for this analysis, where it comes automatically with the CPD.

\subsubsection{Crossed barrel}

\begin{figure}[!ht]
    \centering
    \includegraphics[width = 0.43\textwidth]{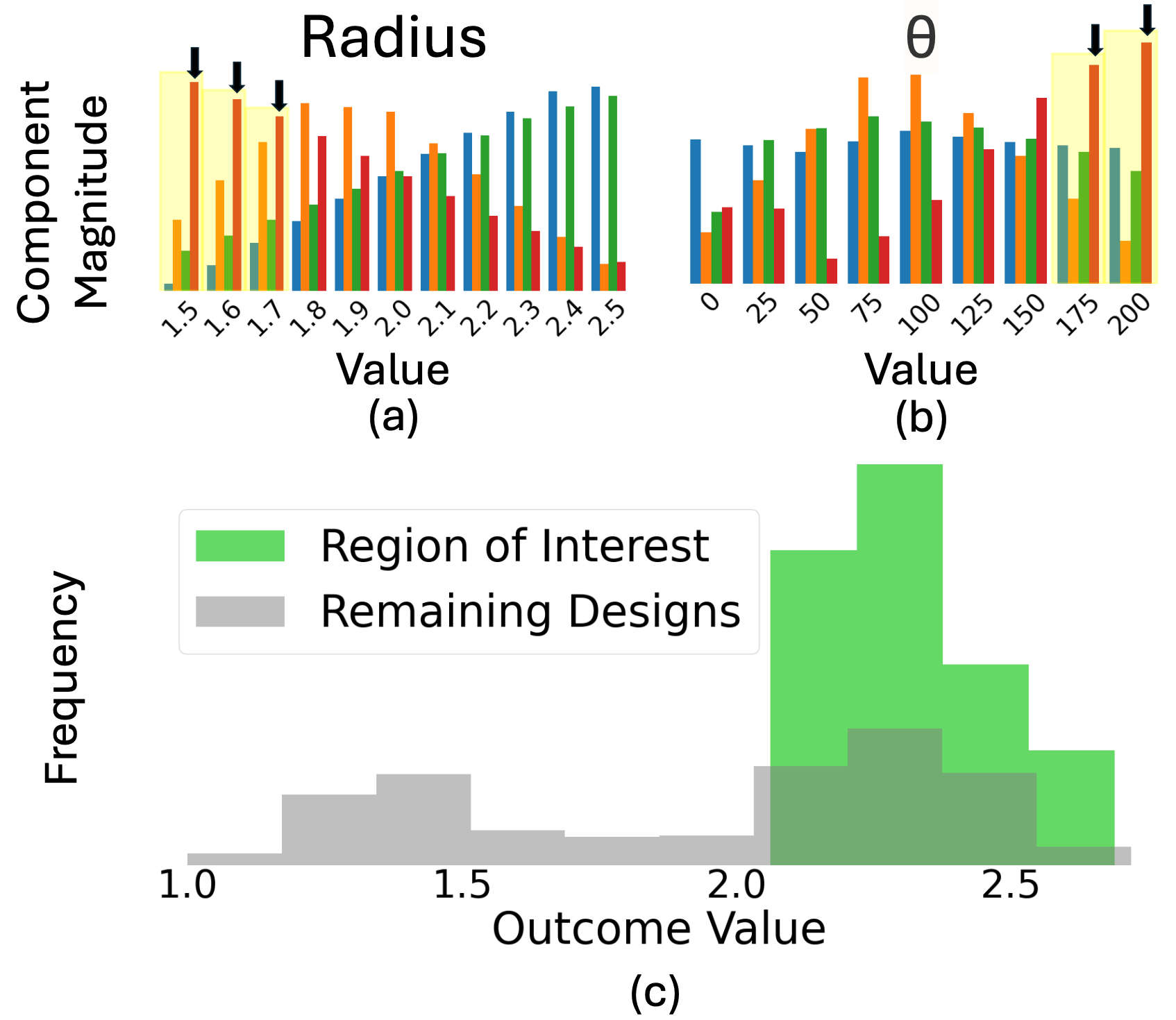}
    \caption{For the crossed barrel dataset, we display the CPD (rank 4) factors. We plot the CPD normalized components for the $\theta$ and radius ($r$) design parameters in (a) and (b), and we can easily identify some interesting patterns. The red colored components (marked with an arrow) have a distinctly increasing trend with high $\theta$ values and with low $r$ values. Interestingly enough, when we look at design parameter values with high expression of this component (highlighted in yellow), we see in the true distribution (c) that these designs have some of the highest toughness values (highlighted green), agreeing with the underlying physics of the problem (see section 3.1.2 discussion). \label{cpd_factors_viz_crossed_barrel}}
\end{figure}

Since the crossed barrel dataset is slightly more complicated, it is hard to analyze any trend in a single design parameter (we could simply do a rank 1 decomposition to analyze each design variable's effect on the outcome, but then prediction performance suffers). For this reason, we try to analyze any clusters of material designs that may be interesting. We use 80\% of the observed data (80\% of 600 observed samples = 480) as input to a CPD rank 4 decomposition, and analyze the normalized factors in Figure \ref{cpd_factors_viz_crossed_barrel}.

The cluster we found in Figure \ref{cpd_factors_viz_crossed_barrel} corresponds to high $\theta$ values and low radius (r) values, which in turn corresponds to high toughness values. This finding agrees with the actual physics of the problem, indicating that we have essentially rediscovered a physical phenomenon as a result of our CPD surrogate model's tensor factors. Having rediscovered this true underlying effect, it is reasonable to assume that experimentalists can use this as a tool to uncover potentially novel patterns in their data.

\textbf{Discussion}: The crossed barrel dataset describes designs of two plates connected by $n$ hollow tubes (with radius $r$ and thickness $t$), which are then twisted at an angle $\theta$. Hollow tubes have traditionally been considered as not only light weight options for structural design but also for high energy absorption applications. Hollow cylinders distribute material on the outside circumference of the structure to maximize the moment of inertia and increase bending resistance. This property is key during quasi-static compression where hollow cylinders of the crossed barrel structure contribute to larger reaction forces under compression resulting in higher energy absorption.

Using our method for interpretation, we discovered a cluster of high-toughness (i.e. high energy absorption) designs, corresponding to high $\theta$ values and low $r$ values, which agrees with the existing literature \cite{duan2021energy, san2023energy, tung2022bio}. While one may think that higher $r$ values would be beneficial, the relationship is not simply monotonic and linear. Interestingly, designs with lower $r$ values, which do contribute to increased bending efficiency, have a higher energy absorption. This is likely due to structures with high $r$ values being brittle and breaking during compression where lower $r$ values have sufficient mass to not fail in a brittle way but can leverage the increase in bending resistance from a hollow circular geometry. To this end, the hollow nature allows for gradual dissipation of energy during compression in addition to the added benefit of bending efficiency. In addition to the positive contributions from hollow struts, the large $\theta$ values play a key role in resulting in superior energy absorption properties because the curvature of the struts allow the conversion of axial loading into a combination of bending, torsion, and compression. In this way, energy is dissipated through other mechanisms and contributes to the plateau region during deformation \cite{duan2021energy, san2023energy, tung2022bio}.  

However, we also note that increasing these properties monotonically do not exclusively result in high energy absorption since as $\theta$ increases and $r$ decreases to their respective maximum values, the crossed barrel structures become either too stiff and violate the force constraints and thus result in structures that are low energy absorption or completely shatter during compression experiments. As a result, high energy absorbing structures require the careful combination of high $\theta$ and low $r$ values (i.e. there exist non-optimal designs with high $\theta$ and/or low $r$, as identified in \cite{gongora2020bayesian}). This also means it is not such a trivial pattern, as we cannot simply monotonically increase $\theta$ and decrease $r$ to their most extreme values, yet we are still able to identify a high-performing cluster using the CPD factors. 

\subsubsection{Cogni-e-Spin}

\begin{figure}[!ht]
    \centering
    \includegraphics[width = 0.43\textwidth]{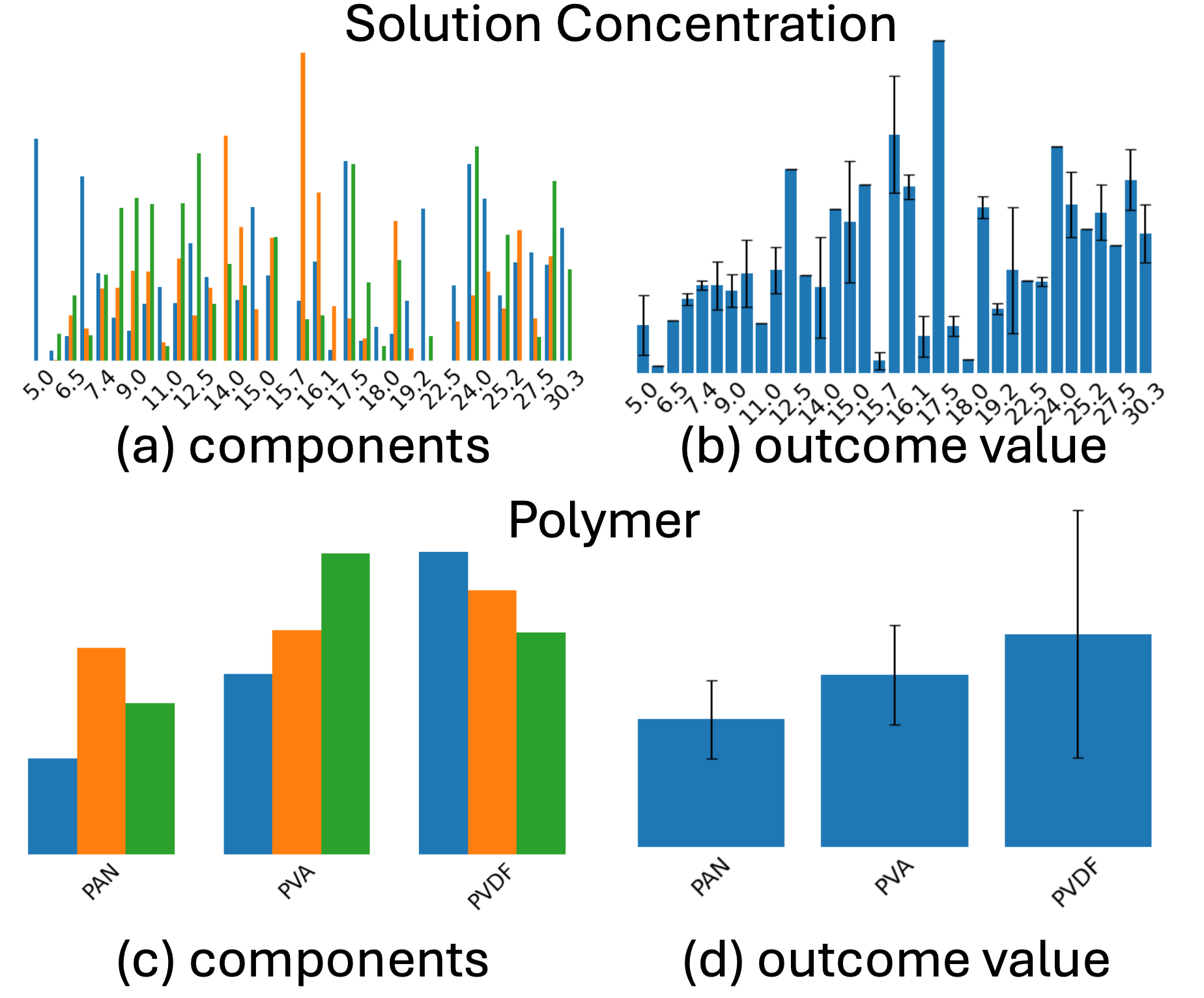}
    \caption{For the Cogni-e-Spin dataset, we display the CPD (rank 3) factors. We can do a very similar thing as with the lattice structures dataset. We can very easily see the effect of each variable on the outcome of interest (in this case fiber diameter). The solution concentration has an extremely varied effect on the outcome in (a) \& (b). We can also see which polymer types tend to have higher fiber diameter in (c) \& (d). \label{cpd_factors_viz_cogni_spin}}
\end{figure}

For the Cogni-e-Spin dataset, we can do a similar analysis as with the lattice dataset in Figure \ref{cpd_factors_viz_cogni_spin}. We can see how the CPD model uses each design parameter for predictions. With the solution concentration parameter we notice the effect is varied across parameter values, as reflected in the component magnitudes in (a) and the true distribution in (b). We can furthermore pickup this trend with the type of polymer and the effect on the outcome (in this case nanofiber diameter) from the components in (c) which correctly represents their outcome values in (d).

\subsection{Uniform Sampling for Training Data}

When looking at model predictions, first we consider a more "traditional" machine learning setting, where we sample the training data uniformly across the entire design space. We use an 80\%-20\% train-test split and show the results in Figure \ref{uniform_sample_results_table}.

\begin{figure*}[!h]
\centering
\small
\setlength{\tabcolsep}{3pt}
\begin{tabular}{r|cccc|cccc|cccc}
\multicolumn{1}{c}{} & \multicolumn{4}{c}{Lattice Dataset} & \multicolumn{4}{c}{Crossed Barrel Dataset} & \multicolumn{4}{c}{Cogni-e-Spin Dataset} \\
\toprule
 & R² & MAE & RMSE & MAPE & R² & MAE & RMSE & MAPE & R² & MAE & RMSE & MAPE \\
\midrule
LR & 0.86 ± 0.0 & 0.06 ± 0.0 & 0.08 ± 0.0 & 0.17 ± 0.1 & 0.34 ± 0.1 & 0.15 ± 0.0 & 0.19 ± 0.0 & 0.42 ± 0.1 & 0.22 ± 0.2 & 0.10 ± 0.0 & 0.14 ± 0.0 & 0.81 ± 0.5 \\
RF & 0.98 ± 0.0 & 0.02 ± 0.0 & 0.03 ± 0.0 & 0.07 ± 0.0 & 0.72 ± 0.1 & \textbf{0.08 ± 0.0} & 0.13 ± 0.0 & \textbf{0.19 ± 0.0} & \underline{0.51 ± 0.1} & \textbf{0.06 ± 0.0} & \underline{0.11 ± 0.0} & 0.56 ± 0.4 \\
XGBoost & \textbf{0.99 ± 0.0} & \textbf{0.01 ± 0.0} & \underline{0.02 ± 0.0} & 0.04 ± 0.0 & 0.64 ± 0.1 & \underline{0.09 ± 0.0} & 0.14 ± 0.0 & \underline{0.21 ± 0.0} & 0.45 ± 0.1 & 0.07 ± 0.0 & 0.12 ± 0.0 & 0.59 ± 0.4 \\
CatBoost & 0.95 ± 0.0 & 0.03 ± 0.0 & 0.05 ± 0.0 & 0.13 ± 0.1 & \textbf{0.76 ± 0.0} & \underline{0.09 ± 0.0} & \textbf{0.12 ± 0.0} & 0.24 ± 0.1 & \textbf{0.55 ± 0.2} & \textbf{0.06 ± 0.0} & \textbf{0.10 ± 0.0} & \underline{0.53 ± 0.3} \\
GP & \textbf{0.99 ± 0.0} & \textbf{0.01 ± 0.0} & \textbf{0.01 ± 0.0} & \textbf{0.02 ± 0.0} & \underline{0.73 ± 0.0} & \underline{0.09 ± 0.0} & \textbf{0.12 ± 0.0} & \underline{0.21 ± 0.0} & 0.22 ± 0.3 & 0.08 ± 0.0 & 0.14 ± 0.0 & 0.62 ± 0.3 \\
MLP & 0.77 ± 0.1 & 0.08 ± 0.0 & 0.10 ± 0.0 & 0.22 ± 0.1 & 0.41 ± 0.1 & 0.15 ± 0.0 & 0.18 ± 0.0 & 0.37 ± 0.1 & 0.11 ± 0.1 & 0.11 ± 0.0 & 0.15 ± 0.0 & 0.91 ± 0.5 \\
\midrule
CPD & \textbf{0.99 ± 0.0} & \textbf{0.01 ± 0.0} & \underline{0.02 ± 0.0} & \underline{0.03 ± 0.0} & 0.68 ± 0.0 & 0.10 ± 0.0 & 0.14 ± 0.0 & 0.23 ± 0.0 & -0.33 ± 0.3 & 0.11 ± 0.0 & 0.18 ± 0.0 & \textbf{0.49 ± 0.2} \\
CPD-S & 0.91 ± 0.0 & 0.05 ± 0.0 & 0.06 ± 0.0 & 0.17 ± 0.1 & 0.67 ± 0.1 & 0.10 ± 0.0 & 0.14 ± 0.0 & 0.22 ± 0.0 & 0.34 ± 0.2 & 0.08 ± 0.0 & 0.13 ± 0.0 & 0.64 ± 0.4 \\
NeAT & 0.98 ± 0.0 & 0.02 ± 0.0 & 0.03 ± 0.0 & 0.05 ± 0.0 & 0.72 ± 0.1 & \underline{0.09 ± 0.0} & 0.13 ± 0.0 & \underline{0.21 ± 0.0} & 0.39 ± 0.1 & 0.08 ± 0.0 & 0.12 ± 0.0 & 0.78 ± 0.5 \\
CoSTCo & 0.90 ± 0.0 & 0.05 ± 0.0 & 0.06 ± 0.0 & 0.17 ± 0.1 & 0.55 ± 0.1 & 0.11 ± 0.0 & 0.16 ± 0.0 & 0.26 ± 0.1 & 0.45 ± 0.1 & 0.08 ± 0.0 & 0.12 ± 0.0 & 0.72 ± 0.5 \\

\bottomrule
\end{tabular}

\caption{Here we evaluate the performance of several ML surrogate models in a typical supervised learning setting (i.e. we train on 80\% of the available data, uniformly sampled). We show the best performance for each task (column) in bold, and the second best performance is underlined. We display the mean and standard deviation across 10 iterations. Please note the outcome values are all normalized to the range [0, 1], so the MAE and RMSE are intentionally without units. \label{uniform_sample_results_table}}
\end{figure*}

From these results, it is hard to definitively say that one single model is the best for these datasets; however, the traditional ML methods (mainly GP and CatBoost) seem to have the best performance in general. Our main takeaway though is that tensor models are definitely comparable to traditional ML models in a typical supervised learning scenario.

\subsection{Non-uniform Sampling for Training Data}

\subsubsection{Experimental setup}

\begin{figure*}[!h]
    \centering
    \subfigure[Lattice dataset]{
    \includegraphics[width = 0.3\textwidth]{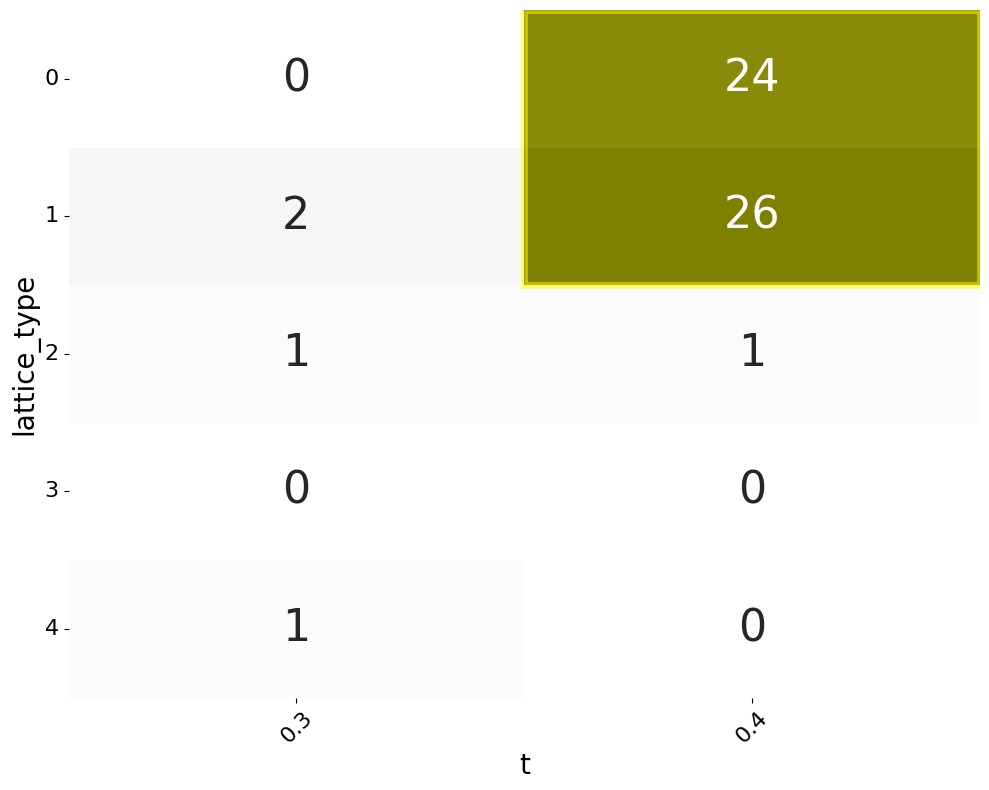}
    }
    \subfigure[Crossed Barrel dataset]{
    \includegraphics[width = 0.3\textwidth]{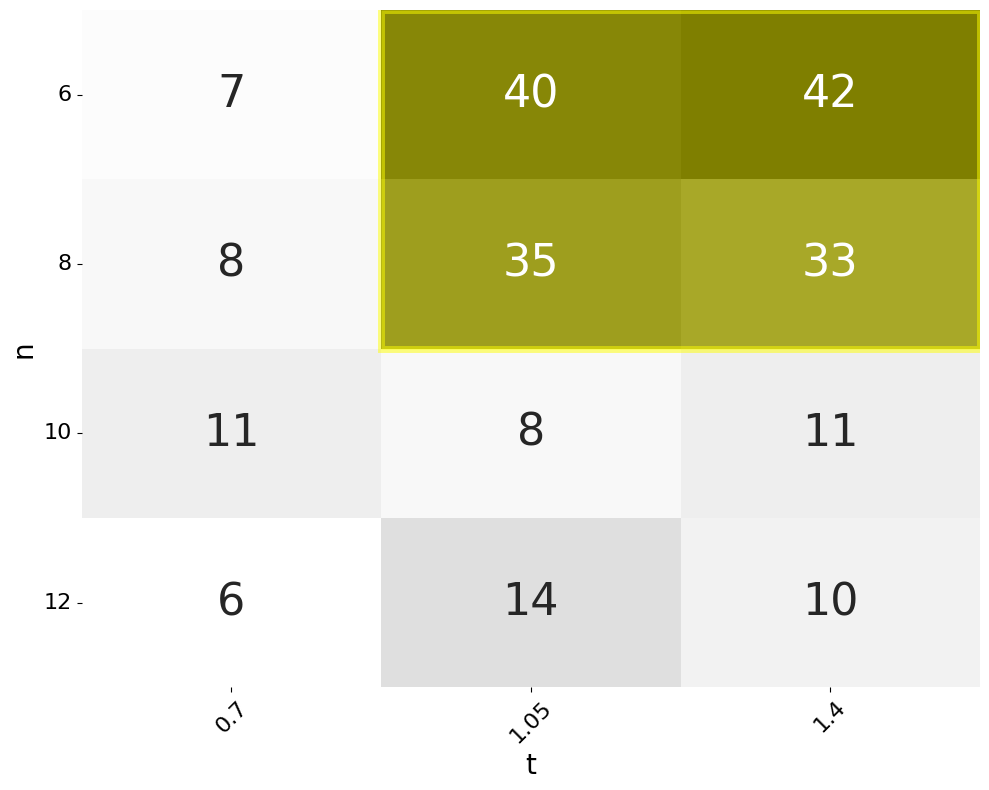}
    }
    \subfigure[Cogni-e-Spin dataset]{
    \includegraphics[width = 0.3\textwidth]{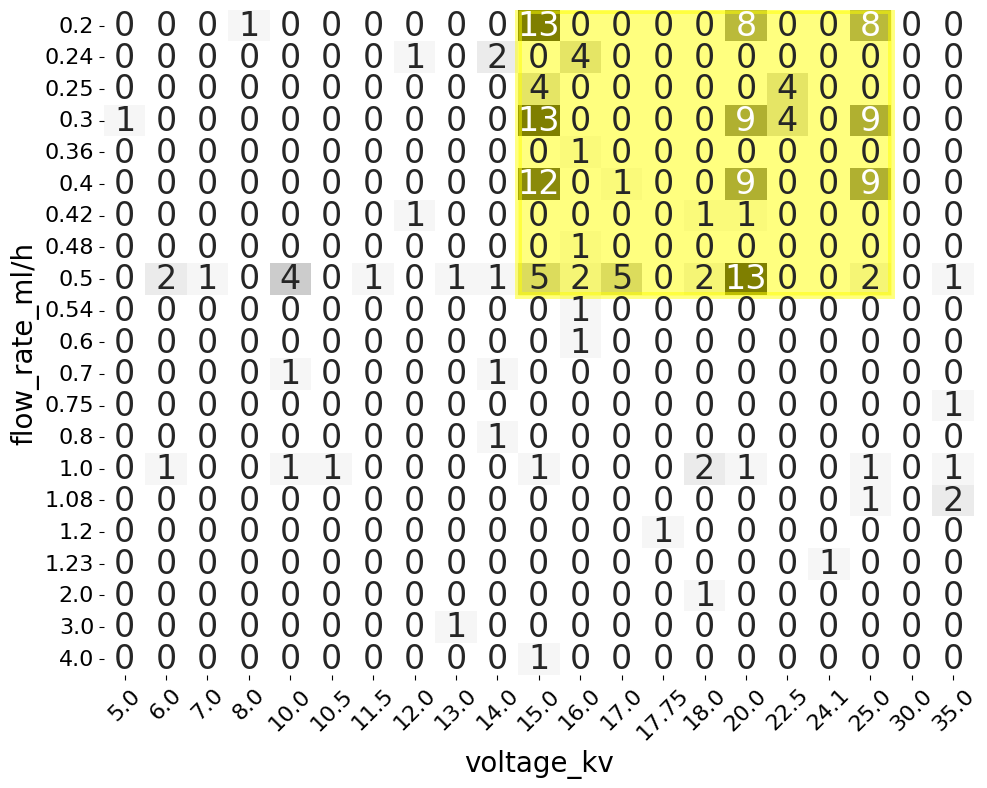}
    }
    \caption{Visualization of our training data for the results in this section, with respect to the 2 design parameters we are sampling in a non-uniform manner (the remaining design parameters are uniformly randomly sampled according to these heatmaps). We sample heavily in the yellow shaded region, and very lightly outside of this region to simulate how an experimentalist might conduct most of their experiments in a specific region of the search space, and not uniformly randomly. \label{biased_sampling_heatmaps}}
\end{figure*}

In practice it is very common to have training data (i.e. experimental/simulation observations) that comes from a non-uniform random sampling of the entire design space. This could be for a number of reasons, including resource constraints (e.g. only generating designs that are cheaper or more convenient). To investigate the behavior of ML methods and tensor methods when the training data comes from a non-uniform sampling of the design space, we sample the training data heavily from a specific region of the design space (with regards to 2 design parameters), and sample significantly less from the remaining part of the space. The exact sampling we use for the experiments is displayed in Figure \ref{biased_sampling_heatmaps}.

\subsubsection{Results}

First we evaluate the aggregate performance of all the methods in Figure \ref{biased_sampling_results_table}.

\begin{figure*}[!h]
\centering
\small
\setlength{\tabcolsep}{3pt}
\begin{tabular}{r|cccc|cccc|cccc}
\multicolumn{1}{c}{} & \multicolumn{4}{c}{Lattice Dataset} & \multicolumn{4}{c}{Crossed Barrel Dataset} & \multicolumn{4}{c}{Cogni-e-Spin Dataset} \\
\toprule
 & R² & MAE & RMSE & MAPE & R² & MAE & RMSE & MAPE & R² & MAE & RMSE & MAPE \\
\midrule
LR & 0.76 ± 0.0 & 0.10 ± 0.0 & 0.13 ± 0.0 & 0.64 ± 0.4 & 0.28 ± 0.0 & 0.18 ± 0.0 & 0.23 ± 0.0 & 1.56 ± 1.4 & 0.24 ± 0.1 & 0.15 ± 0.0 & 0.20 ± 0.0 & 1.38 ± 0.5 \\
RF & 0.77 ± 0.0 & 0.10 ± 0.0 & 0.13 ± 0.0 & 0.57 ± 0.4 & \underline{0.59 ± 0.0} & \textbf{0.12 ± 0.0} & \underline{0.18 ± 0.0} & 0.77 ± 0.6 & \underline{0.38 ± 0.1} & \textbf{0.12 ± 0.0} & \underline{0.18 ± 0.0} & 1.13 ± 0.4 \\
XGBoost & 0.72 ± 0.1 & 0.10 ± 0.0 & 0.14 ± 0.0 & 0.56 ± 0.4 & 0.53 ± 0.0 & \textbf{0.12 ± 0.0} & 0.19 ± 0.0 & \textbf{0.42 ± 0.2} & \underline{0.38 ± 0.1} & \textbf{0.12 ± 0.0} & \underline{0.18 ± 0.0} & \underline{0.95 ± 0.3} \\
CatBoost & 0.67 ± 0.1 & 0.12 ± 0.0 & 0.15 ± 0.0 & 0.84 ± 0.6 & 0.55 ± 0.0 & 0.14 ± 0.0 & \underline{0.18 ± 0.0} & 1.24 ± 1.1 & 0.36 ± 0.1 & 0.13 ± 0.0 & \underline{0.18 ± 0.1} & 1.17 ± 0.4 \\
GP & \underline{0.79 ± 0.1} & \underline{0.09 ± 0.0} & \underline{0.12 ± 0.0} & \underline{0.48 ± 0.3} & \textbf{0.68 ± 0.0} & \textbf{0.12 ± 0.0} & \textbf{0.16 ± 0.0} & \underline{0.73 ± 0.6} & -0.21 ± 0.4 & 0.16 ± 0.1 & 0.25 ± 0.1 & 1.17 ± 0.5 \\
MLP & -0.24 ± 0.9 & 0.22 ± 0.1 & 0.27 ± 0.1 & 0.86 ± 0.5 & 0.15 ± 0.2 & 0.20 ± 0.0 & 0.25 ± 0.0 & 1.57 ± 1.4 & -0.05 ± 0.2 & 0.18 ± 0.0 & 0.23 ± 0.0 & 1.66 ± 0.6 \\
\midrule
CPD & -0.81 ± 0.9 & 0.25 ± 0.1 & 0.34 ± 0.1 & 0.71 ± 0.3 & 0.25 ± 0.1 & 0.17 ± 0.0 & 0.24 ± 0.0 & 1.05 ± 1.0 & -0.79 ± 0.3 & 0.22 ± 0.1 & 0.30 ± 0.1 & \textbf{0.86 ± 0.2} \\
CPD-S & 0.52 ± 0.2 & 0.13 ± 0.0 & 0.17 ± 0.0 & 0.64 ± 0.4 & 0.56 ± 0.1 & 0.14 ± 0.0 & \underline{0.18 ± 0.0} & 0.92 ± 0.9 & -0.02 ± 0.2 & 0.16 ± 0.0 & 0.23 ± 0.1 & 1.28 ± 0.5 \\
NeAT & 0.75 ± 0.1 & 0.10 ± 0.0 & 0.13 ± 0.0 & 0.58 ± 0.3 & 0.50 ± 0.1 & 0.14 ± 0.0 & 0.19 ± 0.0 & 1.04 ± 1.0 & 0.34 ± 0.1 & 0.14 ± 0.0 & \underline{0.18 ± 0.0} & 1.39 ± 0.5 \\
CoSTCo & \textbf{0.84 ± 0.0} & \textbf{0.08 ± 0.0} & \textbf{0.10 ± 0.0} & \textbf{0.45 ± 0.3} & 0.42 ± 0.1 & 0.14 ± 0.0 & 0.21 ± 0.0 & 1.00 ± 1.0 & \textbf{0.43 ± 0.1} & \textbf{0.12 ± 0.0} & \textbf{0.17 ± 0.0} & 1.19 ± 0.4 \\

\bottomrule
\end{tabular}

\caption{Here we evaluate the performance of our ML surrogate models when our training set comes from a biased sampling of the search space. We show the best performance for each task (column) in bold, and the second best performance is underlined. We display the mean and standard deviation across 10 iterations. Please note the outcome values are all normalized to the range [0, 1], so the MAE and RMSE are intentionally without units. We show t-tests for these results in Figure \ref{t_test_table}. \label{biased_sampling_results_table}}

\end{figure*}

We see CoSTCo generally outperforms other methods in the Lattice \& Cogni-e-Spin datasets in this biased sampling scenario, when considering aggregated results. To compare these results we perform t-tests in Figure \ref{t_test_table}.

\begin{figure}[!h]
    \centering
    \begin{tabular}{ccccccccc}
    & \multicolumn{2}{c}{Lattice} & & \multicolumn{2}{c}{Crossed Barrel} & & \multicolumn{2}{c}{Cogni-e-Spin} \\
    \toprule
    Baseline & $t$ & $p$ & & $t$ & $p$ & & $t$ & $p$ \\
    \midrule
    XGBoost & 2.74 & 0.01 & & -3.45 & 0.00 & & 2.02 & 0.06 \\
    CatBoost & 5.90 & 0.00 & & -5.30 & 0.00 & & 0.38 & 0.71 \\
    GP & 1.40 & 0.18 & & -9.60 & 0.00 & & 4.10 & 0.00 \\
    RF & 2.48 & 0.02 & & -5.42 & 0.00 & & 0.44 & 0.67 \\
    \bottomrule
    \end{tabular}
    \caption{We compute $t$-tests for $R^2_T-R^2_B$ where $R^2_T$ is the $R^2$ for the tensor model and $R^2_B$ for the baseline model. We evaluate the $R^2$ over 10 initializations and compare each baseline to the best performing tensor model for that dataset (CPD-S for Crossed Barrel, CoSTCo for remaining).}
    \label{t_test_table}
\end{figure}

\begin{figure}[!ht]
    \centering
    \includegraphics[width = 0.48\textwidth]{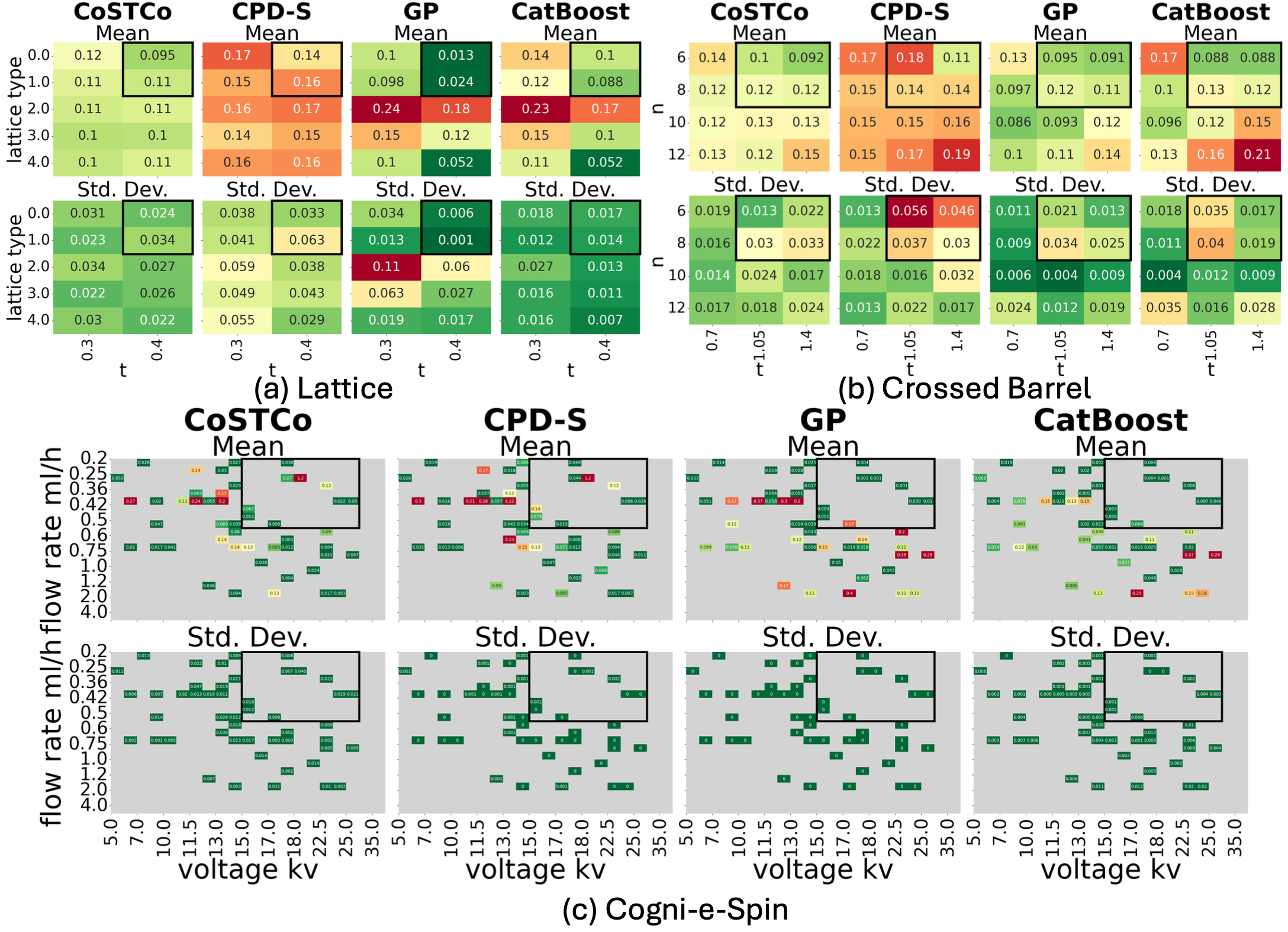}
    \caption{Visualization of model generalization under biased sampling conditions. We display the average and standard deviation of the MAE over 5 iterations for each combination of two design parameters (from Figure \ref{biased_sampling_heatmaps}). The region where we have the most training data is outlined in black. For the Cogni-e-Spin dataset, not all combinations of design parameters are observed, corresponding to gray cells in plot (c). For each dataset we display the two best performing tensor models and the two best performing traditional ML models. An enlarged version of this figure is in the appendix. \label{biased_sampling_error_by_region}}
\end{figure}

We can more closely observe the models' behaviors with the disaggregated errors by subsection of the design space in Figure \ref{biased_sampling_error_by_region}.

Here we are able to more closely observe how the traditional ML methods seem to overfit to the region where the training data is heavily sampled from, in both the Lattice and Cogni-e-Spin datasets, where CoSTCo (and CPD-S for Cogni-e-Spin) still seem to be able to generalize very well. Although the traditional ML methods outperform the tensor methods within the heavily sampled region, tensor methods seem to be able to better generalize outside of this region. We also show the standard deviation, indicating CoSTCo \& CPD-S can give more consistent predictions for the lattice data.

\subsubsection{Performance for different levels of biased sampling}

\begin{figure}[!h]
    \centering
    \includegraphics[width = 0.48\textwidth]{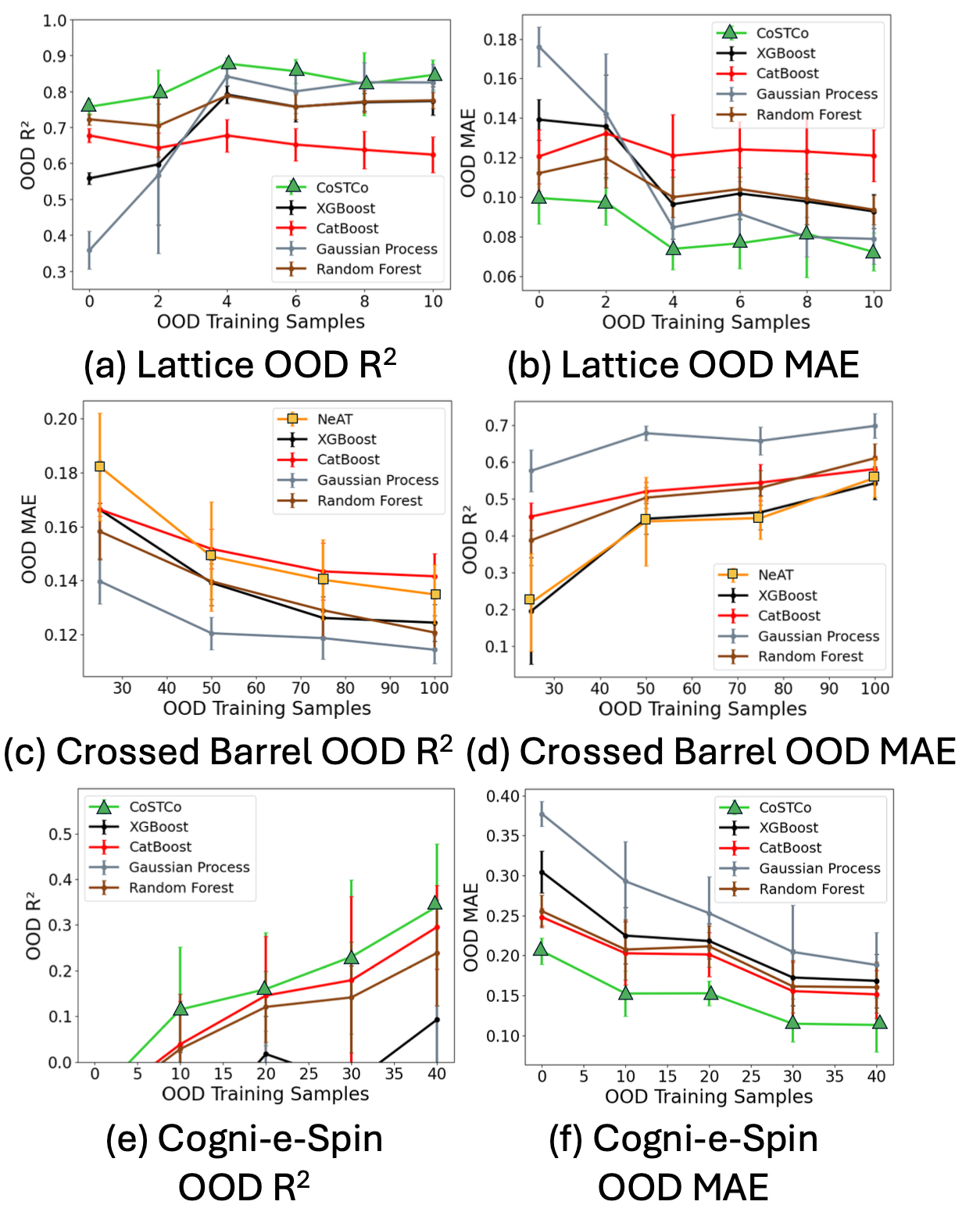}
    \caption{We display the $R^2$ and MAE for the out of distribution (OOD) samples, for various amount of OOD training samples, and a fixed number of in distribution samples (same as Figure \ref{biased_sampling_heatmaps}). We see CoSTCo outperforms the other models with low OOD samples on the Lattice and Cogni-e-Spin data. \label{biased_line_plots}}
\end{figure}

We also want to observe how these models behave as we go from very little to "enough" out of distribution (OOD) samples. We plot these results in Figure \ref{biased_line_plots}, showing the $R^2$ and MAE for \textit{only} the OOD samples, to see how well the models generalize.

In Figure \ref{biased_line_plots} we observe a similar story as before, where we see the excellent generalization capabilities of CoSTCo in the Lattice and Cogni-e-Spin datasets. In the crossed barrel dataset, the traditional ML methods (mainly GP) seem better.
\subsection{Strength of the Low-rank Assumption}

We generally want to study the strength and the limits of our low-rank assumption in these types of material design datasets. In theory, for a truly low-rank dataset we should be able to recover the same or similar tensor factors, and in turn the same or similar predictions, across runs. This should hold regardless of the random initializations, train-test splits, or if the training data is sampled in a biased manner (like our experiments in section 3.3). Here we empirically study to what extent this holds in practice.

\subsubsection{Rank versus Performance}

Perhaps the simplest thing we can do to test the low-rank assumption, is plot the CPD performance for varying rank decompositions.

\begin{figure}[!ht]
    \centering
    \includegraphics[width = 0.45\textwidth]{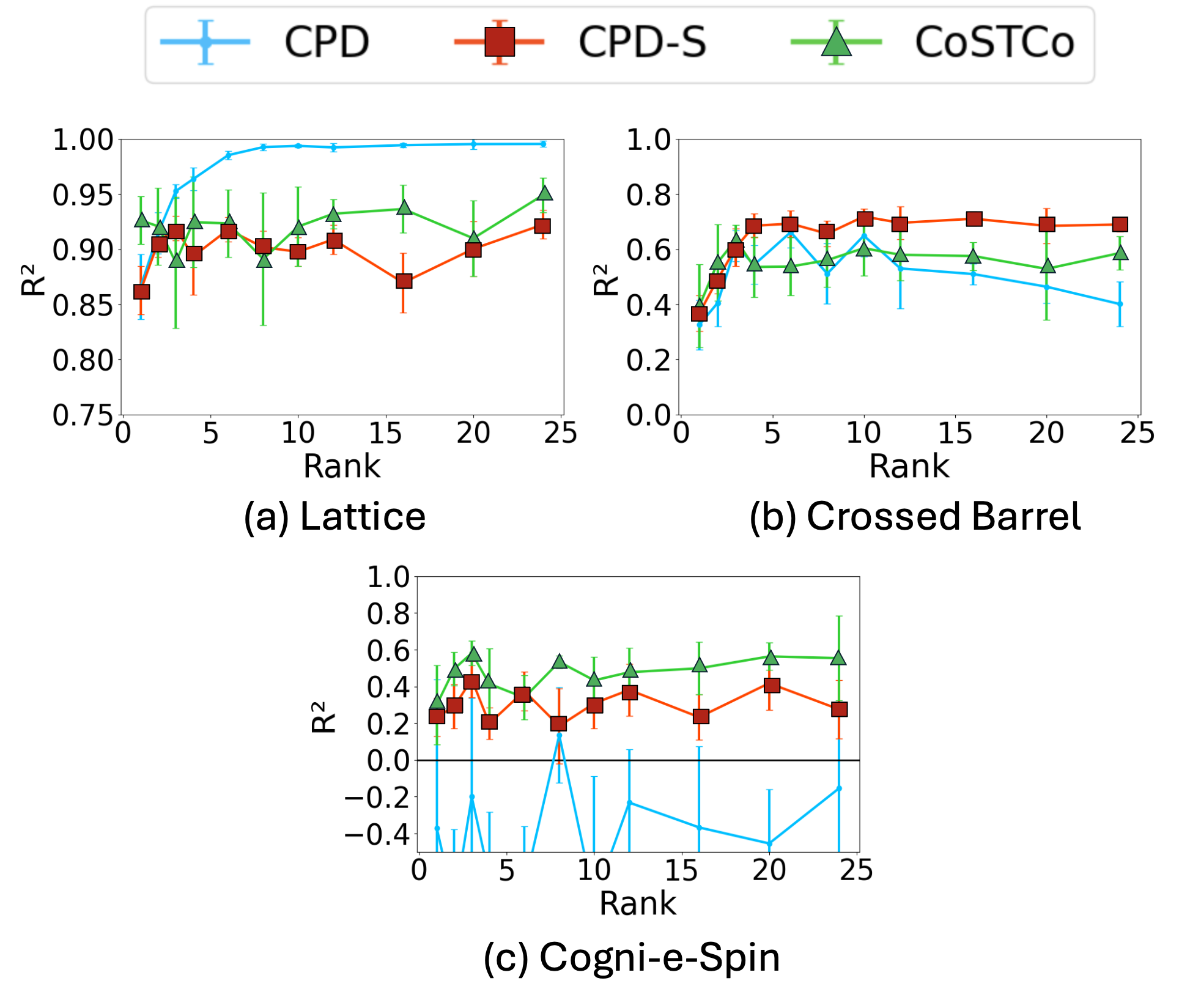}
    \caption{We plot the performance ($R^2$) for different ranks, displaying the mean and standard deviation of 5 iterations. \label{rank_plots}}
\end{figure}

In Figure \ref{rank_plots} we plot the performance ($R^2$) for CPD, CPD-S, and CoSTCo for a variety of rank decompositions, in order to study the robustness to the rank selection. We notice for the lattice dataset, a CPD for a variety of our chosen ranks from 3 to 24 give excellent performance (>0.95 $R^2$), indicating a strong low-rank structure. For the crossed barrel dataset, we get pretty decent performance with the CPD, although there is definitely a sweet spot for the rank, between ranks 4 to 10. We do notice that we can perhaps alleviate some of this performance degradation by imposing a simple smoothness constraint (CPD-S) on the factors to more accurately model the smooth optimization landscape. For the Cogni-e-Spin dataset this effect is more pronounced, as a CPD gets very poor results, indicating a looser adherence to the low-rank assumption, but we can perhaps alleviate some of this with the CPD-S. Finally the CoSTCo model seems very robust to the rank selection (as expected, given its use of neural networks -- reducing the reliance upon the tensor factors).

\subsubsection{Interpretable Factors' Robustness to Random Initializations}

Here we want to study how robust our interpretable tensor factors are to random initializations, because ideally we would be able to recover the same interpretations across experiments. In Figure \ref{random_init_fms_table} we compute CPD tensor factors for 10 different random initializations (train test split, factor initialization, and both) and compute the pairwise FMS values (skipping duplicate comparisons) to see how similar the tensor factors.

\begin{figure}[!h]
    \centering
    \begin{tabular}{cccc}
     & Train Split & Factors & Both \\
    \toprule
    Lattice & 0.99 ± 0.0 & 0.83 ± 0.1 & 0.89 ± 0.1 \\
    Crossed Barrel & 0.91 ± 0.1 & 0.66 ± 0.1 & 0.63 ± 0.1 \\
    Cogni-e-Spin & 0.64 ± 0.1 & 0.42 ± 0.1 & 0.36 ± 0.1 \\
    \bottomrule
    \end{tabular}
    \caption{Factor match score (FMS) for CPD factors with different random initializations for the train-test split, tensor factors, and both. For 10 different random initializations we compute one CPD each, and do a pairwise comparison of FMS scores (excluding duplicate comparisons).}
    \label{random_init_fms_table}
\end{figure}

From Figure \ref{random_init_fms_table} we notice that with lattice dataset, the FMS values are very high, again indicating a strong low-rank structure as we are able to reliably recover the optimal factors. In the crossed barrel dataset we get decent FMS scores, though not as strong as with the lattice data. For the Cogni-e-Spin data, we get very poor FMS scores, indicating a weaker low-rank structure.

\subsubsection{Interpretable Factors' Robustness to Non-uniform Sampling}

We know we can achieve better generalizability with CoSTCo for non-uniformly sampled training data, but here we want to see if we can still interpret the factors of a CPD in this scenario. For each dataset we compute a CPD (rank 3), one with a 80\%-20\% uniformly sampled train-test split, and one using the biased training data (refer to Figure \ref{biased_sampling_heatmaps}). In a simple effort to increase reliability, we use 5 different random initializations and take the best (in terms of training loss). Then we compute the factor match score (FMS) between the two sets of factors and display the results in Figure \ref{fms_table}. We assume that since we were already able to interpret the CPD factors earlier, that if we can generate similar tensor factors to this (in terms of FMS), but using a non-uniform sampling, then we can interpret these factors as well. We see in Figure \ref{fms_table} that for the Lattice and Crossed Barrel datasets, we get decently high FMS scores (0.72 and 0.65, respectively), indicating that even using a non-uniform sampling of the design space, we can generate interpretable tensor factors using a CPD. The Cogni-e-Spin dataset, however, has a low FMS score of 0.38.

\begin{figure}[!h]
    \centering
    \begin{tabular}{cccc}
    & Lattice & Crossed Barrel & Cogni-e-Spin \\
    \toprule
    FMS & 0.72 ± 0.04 & 0.65 ± 0.06 & 0.38 ± 0.04 \\
    \bottomrule
    \end{tabular}
    \caption{Factor match score (FMS) for CPD (rank 3) factors for uniform \& non-uniform sampled train data. We display the average and standard deviation over 5 iterations.}
    \label{fms_table}
\end{figure}

\section{Conclusion}

Overall we study two common issues with ML-based surrogate modeling for material design: (i) interpretability and (ii) generalization when the data is from a non-uniform sampling.

We suggest the use of tensor methods as a way to overcome challenges in interpretable surrogate modeling for material design, as well as providing a robust ML model when the training data comes from a non-uniform sampling of the entire design space. We perform a rigorous evaluation of tensor methods compared with traditional ML methods in a variety of datasets and scenarios. We observe different settings where ML methods are superior (e.g. uniformly sampled training data), and also settings where tensor methods may be desirable (e.g. interpretability, biased sampling).
\section{Limitations and Ethical Considerations}

One limitation of this work is the inability to consider how these methods behave as the design parameters take more values. Due to the expensive nature of building these datasets (e.g., a gridsearch of experimentally validated materials), the dataset size is limited. The low-rank assumption is a possible limitation. If there is, however, a low-rank structure in the data, it allows for an interpretable CPD model to infer missing entries very effectively. Unfortunately, computing the rank of a tensor is NP-complete \cite{haastad1990tensor}, but there are algorithms for approximation \cite{shiao2024frappe}. Another important aspect is the discretization of continuous design parameters. Unfortunately, the datasets we use in this study are given to us discretized, so we cannot study this for these material design problems, but there is work on finding the optimal discretization for tensor analysis \cite{pasricha2023harvester}.

There's no concern for data privacy, consent, or potential misuse.

\begin{acks}
Research was supported by National Science Foundation CAREER grant no. IIS 2046086  and CREST Center for Multidisciplinary Research Excellence in CyberPhysical Infrastructure Systems (MECIS) grant no. 2112650. LLNL R\&R IM Number: LLNL-JRNL-2020789.
\end{acks}

\section*{Generative AI Disclosure}

Generative AI (ChatGPT) was only used to help with writing the code for some plots/tables in the results sections.

\clearpage

\bibliographystyle{ACM-Reference-Format}
\bibliography{refs}

@article{gongora2024accelerating,
  title={Accelerating the design of lattice structures using machine learning},
  author={Gongora, Aldair E and Friedman, Caleb and Newton, Deirdre K and Yee, Timothy D and Doorenbos, Zachary and Giera, Brian and Duoss, Eric B and Han, Thomas Y-J and Sullivan, Kyle and Rodriguez, Jennifer N},
  journal={Scientific Reports},
  volume={14},
  number={1},
  pages={13703},
  year={2024},
  publisher={Nature Publishing Group UK London}
}

@inproceedings{ahn2021time,
  title={Time-aware tensor decomposition for sparse tensors},
  author={Ahn, Dawon and Jang, Jun-Gi and Kang, U},
  booktitle={2021 IEEE 8th International Conference on Data Science and Advanced Analytics (DSAA)},
  pages={1--2},
  year={2021},
  organization={IEEE}
}

@inproceedings{liu2019costco,
  title={Costco: A neural tensor completion model for sparse tensors},
  author={Liu, Hanpeng and Li, Yaguang and Tsang, Michael and Liu, Yan},
  booktitle={Proceedings of the 25th ACM SIGKDD International Conference on Knowledge Discovery \& Data Mining},
  pages={324--334},
  year={2019}
}

@inproceedings{balazevic2019tucker,
title={TuckER: Tensor Factorization for Knowledge Graph Completion},
author={Bala\v{z}evi\'c, Ivana and Allen, Carl and Hospedales, Timothy M},
booktitle={Empirical Methods in Natural Language Processing},
year={2019}
}

@article{kolda2009tensor,
  title={Tensor decompositions and applications},
  author={Kolda, T.G. and Bader, B.W.},
  journal={SIAM review},
  volume={51},
  number={3},
  year={2009},
  publisher={Citeseer}
}

@article{sidiropoulos2016tensor,
  title={Tensor Decomposition for Signal Processing and Machine Learning},
  author={Sidiropoulos, Nicholas D and De Lathauwer, Lieven and Fu, Xiao and Huang, Kejun and Papalexakis, Evangelos E and Faloutsos, Christos},
  journal={IEEE Signal Processing Magazine},
  issue_date = {(to appear)},
  year = 2016,    
  publisher = {IEEE}
}

@INPROCEEDINGS{10825934,
  author={Pakala, Shaan and Graw, Bryce and Ahn, Dawon and Dinh, Tam and Mahin, Mehnaz Tabassum and Tsotras, Vassilis and Chen, Jia and Papalexakis, Evangelos E.},
  booktitle={2024 IEEE International Conference on Big Data (BigData)}, 
  title={Automating Data Science Pipelines with Tensor Completion}, 
  year={2024},
  volume={},
  number={},
  pages={1075-1084},
  doi={10.1109/BigData62323.2024.10825934}
}

@inproceedings{ahnneural,
  title={NEURAL ADDITIVE TENSOR DECOMPOSITION FOR SPARSE TENSORS},
  author={Ahn, Dawon and Saini, Uday Singh and Papalexakis, Evangelos E and Payani, Ali},
  booktitle={33rd ACM International Conference on Information and Knowledge Management},
 year={2024},
 organization={ACM}
}

@article{kingma2014adam,
  title={Adam: A method for stochastic optimization},
  author={Kingma, Diederik P},
  journal={arXiv preprint arXiv:1412.6980},
  year={2014}
}

@article{pakala2025tensor,
  title={Tensor Completion for Surrogate Modeling of Material Property Prediction},
  author={Pakala, Shaan and Ahn, Dawon and Papalexakis, Evangelos},
  journal={arXiv preprint arXiv:2501.18137},
  year={2025}
}

@article{liang2021benchmarking,
  title={Benchmarking the performance of Bayesian optimization across multiple experimental materials science domains},
  author={Liang, Qiaohao and Gongora, Aldair E and Ren, Zekun and Tiihonen, Armi and Liu, Zhe and Sun, Shijing and Deneault, James R and Bash, Daniil and Mekki-Berrada, Flore and Khan, Saif A and others},
  journal={npj Computational Materials},
  volume={7},
  number={1},
  pages={188},
  year={2021},
  publisher={Nature Publishing Group UK London}
}

@article{mahdian2026cogni,
  title={Cogni-e-SpinDB 1.0: Open Dataset of Electrospinning Parameter Configurations and Resultant Nanofiber Morphologies},
  author={Mahdian, Mehrab and Stummer, Tamas and Sepsik, Norman and Ender, Ferenc and Balogh-Weiser, Diana and Pardy, Tamas},
  journal={Scientific Data},
  year={2026},
  publisher={Nature Publishing Group UK London}
}

@incollection{NIPS2017_7062,
title = {A Unified Approach to Interpreting Model Predictions},
author = {Lundberg, Scott M and Lee, Su-In},
year = {2017},
publisher = {Curran Associates, Inc.},
url = {http://papers.nips.cc/paper/7062-a-unified-approach-to-interpreting-model-predictions.pdf}
}

@article{gongora2020bayesian,
  title={A Bayesian experimental autonomous researcher for mechanical design},
  author={Gongora, Aldair E and Xu, Bowen and Perry, Wyatt and Okoye, Chika and Riley, Patrick and Reyes, Kristofer G and Morgan, Elise F and Brown, Keith A},
  journal={Science advances},
  volume={6},
  number={15},
  pages={eaaz1708},
  year={2020},
  publisher={American Association for the Advancement of Science}
}

@article{alderete2022machine,
  title={Machine learning assisted design of shape-programmable 3D kirigami metamaterials},
  author={Alderete, Nicolas A and Pathak, Nibir and Espinosa, Horacio D},
  journal={npj Computational Materials},
  volume={8},
  number={1},
  pages={191},
  year={2022},
  publisher={Nature Publishing Group UK London}
}

@article{challapalli2021machine,
  title={Machine learning assisted design of new lattice core for sandwich structures with superior load carrying capacity},
  author={Challapalli, Adithya and Li, Guoqiang},
  journal={Scientific reports},
  volume={11},
  number={1},
  pages={18552},
  year={2021},
  publisher={Nature Publishing Group UK London}
}

@article{wu2021topology,
  title={Topology optimization of multi-scale structures: a review},
  author={Wu, Jun and Sigmund, Ole and Groen, Jeroen P},
  journal={Structural and Multidisciplinary Optimization},
  volume={63},
  number={3},
  pages={1455--1480},
  year={2021},
  publisher={Springer}
}

@article{jiao2021artificial,
  title={Artificial intelligence-enabled smart mechanical metamaterials: advent and future trends},
  author={Jiao, Pengcheng and Alavi, Amir H},
  journal={International Materials Reviews},
  volume={66},
  number={6},
  pages={365--393},
  year={2021},
  publisher={SAGE Publications Sage UK: London, England}
}

@article{yeo2018materials,
  title={Materials-by-design: computation, synthesis, and characterization from atoms to structures},
  author={Yeo, Jingjie and Jung, Gang Seob and Mart{\'\i}n-Mart{\'\i}nez, Francisco J and Ling, Shengjie and Gu, Grace X and Qin, Zhao and Buehler, Markus J},
  journal={Physica scripta},
  volume={93},
  number={5},
  pages={053003},
  year={2018},
  publisher={IOP Publishing}
}

@article{pan2020design,
  title={Design and optimization of lattice structures: A review},
  author={Pan, Chen and Han, Yafeng and Lu, Jiping},
  journal={Applied Sciences},
  volume={10},
  number={18},
  pages={6374},
  year={2020},
  publisher={MDPI}
}

@article{pedregosa2011scikit,
  title={Scikit-learn: Machine learning in Python},
  author={Pedregosa, Fabian and Varoquaux, Ga{\"e}l and Gramfort, Alexandre and Michel, Vincent and Thirion, Bertrand and Grisel, Olivier and Blondel, Mathieu and Prettenhofer, Peter and Weiss, Ron and Dubourg, Vincent and others},
  journal={the Journal of machine Learning research},
  volume={12},
  pages={2825--2830},
  year={2011},
  publisher={JMLR. org}
}

@article{chen2016xgboost,
  title={XGBoost: A Scalable Tree Boosting System},
  author={Chen, Tianqi},
  journal={Cornell University},
  year={2016}
}

@article{gongora2021using,
  title={Using simulation to accelerate autonomous experimentation: A case study using mechanics},
  author={Gongora, Aldair E and Snapp, Kelsey L and Whiting, Emily and Riley, Patrick and Reyes, Kristofer G and Morgan, Elise F and Brown, Keith A},
  journal={Iscience},
  volume={24},
  number={4},
  year={2021},
  publisher={Elsevier}
}

@article{dorogush2018catboost,
  title={CatBoost: gradient boosting with categorical features support},
  author={Dorogush, Anna Veronika and Ershov, Vasily and Gulin, Andrey},
  journal={arXiv preprint arXiv:1810.11363},
  year={2018}
}

@article{pakala2025surrogate,
  title={Surrogate Modeling for the Design of Optimal Lattice Structures using Tensor Completion},
  author={Pakala, Shaan and Gongora, Aldair E and Giera, Brian and Papalexakis, Evangelos E},
  journal={arXiv preprint arXiv:2510.07474},
  year={2025}
}

@article{koren2009matrix,
  title={Matrix factorization techniques for recommender systems},
  author={Koren, Yehuda and Bell, Robert and Volinsky, Chris},
  journal={Computer},
  volume={42},
  number={8},
  pages={30--37},
  year={2009},
  publisher={IEEE}
}

@article{zhang2026machine,
  title={Machine learning for phase prediction of high entropy carbide ceramics from imbalanced data},
  author={Zhang, Xuemeng and Sun, Jia and Zhang, Yuyu and Fan, Kaifei and Zhang, Zhixiang and Zhang, Yujia and Wu, Keke and Feldmann, Laura and Wu, Lianwei and Riedel, Ralf and others},
  journal={npj Computational Materials},
  year={2026},
  publisher={Nature Publishing Group UK London}
}

@article{sreedev2025leveraging,
  title={Leveraging Machine Learning for Thermoelectric Material Design: Addressing Composition--Property Relations and Data Imbalance Challenges},
  author={Sreedev, Deepa and Kalarikkal, Nandakumar and Kurukkal Balakrishnan, Subila},
  journal={ACS Applied Energy Materials},
  volume={8},
  number={22},
  pages={16934--16946},
  year={2025},
  publisher={ACS Publications}
}

@article{jiang2025review,
  title={A review of machine learning methods for imbalanced data challenges in chemistry},
  author={Jiang, Jian and Zhang, Chunhuan and Ke, Lu and Hayes, Nicole and Zhu, Yueying and Qiu, Huahai and Zhang, Bengong and Zhou, Tianshou and Wei, Guo-Wei},
  journal={Chemical Science},
  year={2025},
  publisher={Royal Society of Chemistry}
}

@article{snapp2023autonomous,
  title={Autonomous discovery of tough structures},
  author={Snapp, Kelsey L and Verdier, Benjamin and Gongora, Aldair and Silverman, Samuel and Adesiji, Adedire D and Morgan, Elise F and Lawton, Timothy J and Whiting, Emily and Brown, Keith A},
  journal={arXiv preprint arXiv:2308.02315},
  year={2023}
}

@article{wang2021surrogate,
  title={Surrogate model via artificial intelligence method for accelerating screening materials and performance prediction},
  author={Wang, Tian and Shao, Mingqi and Guo, Rong and Tao, Fei and Zhang, Gang and Snoussi, Hichem and Tang, Xingling},
  journal={Advanced Functional Materials},
  volume={31},
  number={8},
  pages={2006245},
  year={2021},
  publisher={Wiley Online Library}
}

@article{haastad1990tensor,
  title={Tensor rank is NP-complete},
  author={H{\aa}stad, Johan},
  journal={Journal of algorithms},
  volume={11},
  number={4},
  pages={644--654},
  year={1990},
  publisher={Academic Press, Inc. Orlando, FL, USA}
}

@article{shiao2024frappe,
  title={FRAPPE: fast rank approximation with explainable features for tensors},
  author={Shiao, William and Papalexakis, Evangelos E},
  journal={Data Mining and Knowledge Discovery},
  volume={38},
  number={6},
  pages={4217--4232},
  year={2024},
  publisher={Springer}
}

@article{kossaifi2019tensorly,
  title={Tensorly: Tensor learning in python},
  author={Kossaifi, Jean and Panagakis, Yannis and Anandkumar, Anima and Pantic, Maja},
  journal={Journal of Machine Learning Research},
  volume={20},
  number={26},
  pages={1--6},
  year={2019}
}

@article{duan2021energy,
  title={Energy-absorbing characteristics of hollow-cylindrical hierarchical honeycomb composite tubes inspired a beetle forewing},
  author={Duan, Y and Zhang, T and Zhou, J and Xiao, H and Chen, X and Al Teneiji, M and Guan, ZW and Cantwell, WJ},
  journal={Composite Structures},
  volume={278},
  pages={114637},
  year={2021},
  publisher={Elsevier}
}

@article{san2023energy,
  title={Energy absorption characteristics of bio-inspired hierarchical multi-cell bi-tubular tubes},
  author={San Ha, Ngoc and Pham, Thong M and Chen, Wensu and Hao, Hong},
  journal={International Journal of Mechanical Sciences},
  volume={251},
  pages={108260},
  year={2023},
  publisher={Elsevier}
}

@article{tung2022bio,
  title={Bio-inspired, helically oriented tubular structures with tunable deformability and energy absorption performance under compression},
  author={Tung, Cheng-Che and Chen, Yen-Shuo and Chen, Wen-Fei and Chen, Po-Yu},
  journal={Materials \& Design},
  volume={222},
  pages={111076},
  year={2022},
  publisher={Elsevier}
}

@inproceedings{pasricha2023harvester,
  title={Harvester: Principled Factorization-based Temporal Tensor Granularity Estimation},
  author={Pasricha, Ravdeep S and Saini, Uday Singh and Sidiropoulos, Nicholas D and Fang, Fei and Chan, Kevin and Papalexakis, Evangelos E},
  booktitle={Proceedings of the 2023 SIAM International Conference on Data Mining (SDM)},
  pages={82--90},
  year={2023},
  organization={SIAM}
}

\appendix

\begin{figure*}[!ht]
    \centering
    \includegraphics[width = \textwidth]{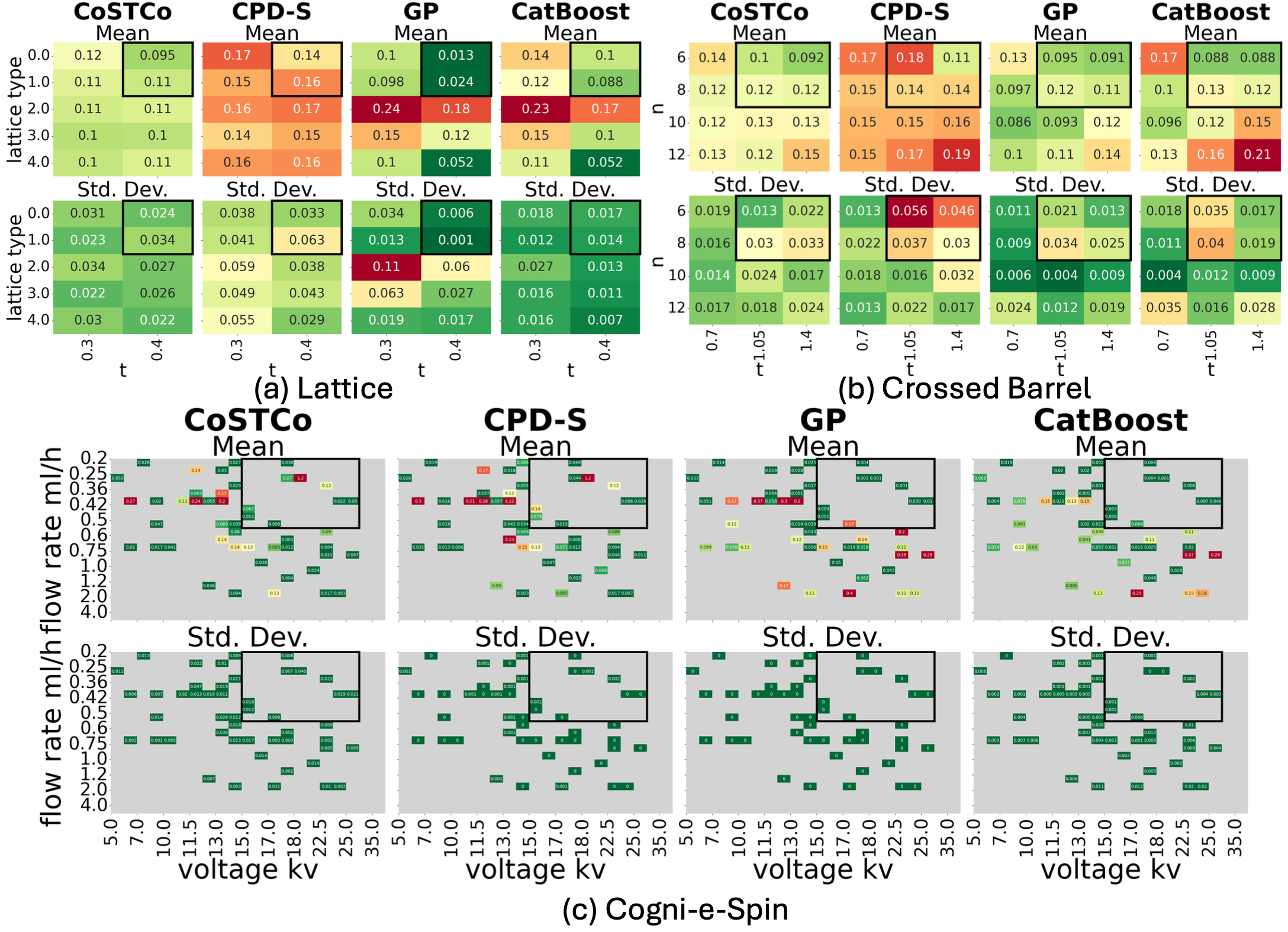}
    \caption{Enlarged version of Figure \ref{biased_sampling_error_by_region}.}
\end{figure*}

\end{document}